\documentclass[preprint,12pt]{elsarticle}

\usepackage{lineno,hyperref}
\modulolinenumbers[5]

\usepackage{tikz}
\usepackage{svg}
\usepackage{multirow}
\usepackage{booktabs}
\usepackage{tabularx}
\usepackage{subcaption}
\usepackage{amsmath,amssymb,amsfonts}
\usepackage{algorithmic}
\usepackage{graphicx}
\usepackage{textcomp}
\usepackage{xcolor}
\usepackage{url}

\begin{document}

\begin{frontmatter}


\title{Transfer Learning for the Efficient Detection of COVID-19 from Smartphone Audio Data}

\author[1]{Mattia Giovanni Campana\corref{cor1}}
\ead{m.campana@iit.cnr.it}
\author[1]{Franca Delmastro}
\ead{f.delmastro@iit.cnr.it}
\author[2]{Elena Pagani}
\ead{elena.pagani@unimi.it}

\cortext[cor1]{Corresponding author}

\affiliation[1]{
    organization={Institute for Informatics and Telematics of the National Research Council of Italy (IIT-CNR)},
    city={Pisa},
    country={Italy}
}

\affiliation[2]{
    organization={Computer Science Department, University of Milano},
    city={Milan},
    country={Italy}
}

\begin{abstract}

Disease detection from smartphone data represents an open research challenge in mobile health (m-health) systems. COVID-19 and its respiratory symptoms are an important case study in this area and their early detection is a potential real instrument to counteract the pandemic situation. The efficacy of this solution mainly depends on the performances of AI algorithms applied to the collected data and their possible implementation directly on the users' mobile devices. Considering these issues, and the limited amount of available data, in this paper we present the experimental evaluation of 3 different deep learning models, compared also with hand-crafted features,  and of two main approaches of transfer learning in the considered scenario: both feature extraction and fine-tuning. Specifically, we considered VGGish, YAMNET, and  L\textsuperscript{3}-Net (including 12 different configurations) evaluated through user-independent experiments on 4 different datasets (13,447 samples in total). Results clearly show the advantages of L\textsuperscript{3}-Net in all the experimental settings as it overcomes the other solutions by 12.3\% in terms of Precision-Recall AUC as features extractor, and by 10\% when the model is fine-tuned.
Moreover, we note that to fine-tune only the fully-connected layers of the pre-trained models generally leads to worse performances, with an average drop of 6.6\% with respect to feature extraction. 
Finally, we evaluate the memory footprints of the different models for their possible applications on commercial mobile devices. 

\end{abstract}



\end{frontmatter}

\section{Introduction}
COVID-19 pandemic largely changed our lives, not only for the impact of the disease on the entire world population, but also for an emerging perception of human weaknesses to the virus's spread and the dependency on the limitations of national healthcare systems. 
In the last couple of years, people became more conscious of the risks, and they increased their trust in digital solutions aimed at counteracting and mitigate the pandemic.  
The digital solutions proposed by researchers are mainly characterized by a massive use of Artificial Intelligence (AI) technologies and big data~\cite{10.1145/3465398, 9141265}; data derived from clinical analysis (e.g., blood test~\cite{Wu2020.04.02.20051136}, X-ray and lung CT images~\cite{gozes2020rapid}), but also from commercial personal and wearable devices collecting physiological and/or audio signals~\cite{exploringcovid2020}.

Mobile health (m-health) systems have the potentiality to become scalable and low-cost solutions for fast screening, aimed at recognizing the onset of new COVID-19 cases, thus possibly preventing new outbreaks.
To this aim, they have to rely on smartphone-embedded sensors, especially microphones, to recognize or distinguish the main COVID symptoms, like cough and breath, from those of either other pathologies or healthy conditions. We investigated previous works in the literature, and we found interesting preliminary works.
Schuller et al.~\cite{10.3389/fdgth.2021.564906} have been the first to investigate how the automatic analysis of speech and audio data can contribute to fight the pandemic crisis, presenting the potential of Computer Audition techniques (CA, i.e., computer-based speech and sound analysis)~\cite{10.3389/fdgth.2020.00005}.
Subsequently, researchers investigated the effective applicability of those techniques in real scenarios.
The main difficulty they encountered was the collection of objective data from large populations.
Initial studies focused on small patients' cohorts~\cite{IMRAN2020100378} even though the target of the work would require a huge amount of data that could not be collected rapidly.
The authors of~\cite{IMRAN2020100378} presented both a preliminary evaluation of a cough detector system aimed at distinguishing cough signals from noise, and an AI tool for COVID-19 diagnosis based on data collected from 70 subjects in controlled environments.
Other works released mobile and web apps to directly collect crowdsourced datasets from the population~\cite{9414576, coswarads, subirana2020hi}. Some of them focused only on  cough audio data, while others introduced also speech signals, trying to improve the classification performances.
Respiratory sound samples (e.g., cough and breath) are generally processed by using standard modeling procedures proposed in the CA literature to extract different sets of features (referred as \emph{hand-crafted acoustic features})~\cite{han2020early}.
Different Deep Learning (DL) approaches  have been proposed~\cite{9208795, ensemble}, including the use of pre-trained models for latent features extraction from raw audio signals, also called \emph{deep audio embeddings}, to enrich standard CA features~\cite{9414576}.


In our previous paper~\cite{9821076}, we investigated the feasibility of using a novel embedding model called \emph{Look, Listen and Learn} (L\textsuperscript{3}-Net)~\cite{Arandjelovic_2017_ICCV} to improve the detection of COVID-19 from respiratory sounds data, including both coughing and breathing recordings.
Specifically, we employed a pre-trained version of L\textsuperscript{3}-Net to extract latent features from audio files, thus relying on transfer learning to characterize raw audio samples in a low-dimensional space, which highlights the differences among the data.
By comparing L\textsuperscript{3}-Net performances with the other proposed models, we demonstrated that it overcomes the others in several standard metrics, scoring an overall gain of 28.57\% of AUC, 23.75\% of Precision, and 39.43\% in terms of Recall.

In this work, we present a more thorough investigation of possible solutions to efficiently detect COVID-19, comparing the use of hand-crafted acoustic features and two different transfer learning techniques aimed at adapting pre-trained DL models: \emph{feature extraction} and \emph{fine-tuning}.
Specifically, we can summarize the novel contributions of this work as follows:

\begin{itemize}
    \item We evaluate the effectiveness of detecting COVID-19 from respiratory sounds data by using both hand-crafted acoustic features and deep audio embeddings extracted by 3 state-of-the-art pre-trained DL models: VGGish, YAMNET, and L\textsuperscript{3}-Net;

    \item We study in details L\textsuperscript{3}-Net, evaluating its classification performances based on different model's configurations;

    \item We compare the two main approaches of transfer learning for model adaptation, i.e., features extraction and fine-tuning, identifying their advantages and drawbacks in our application scenario;

    \item We perform an extensive evaluation of the considered solutions through a series of user-independent experiments by using 4 different benchmark datasets, which correspond to 8,726 data samples more than those we used in our previous work;

    \item Finally, we study the memory footprint of both transfer learning approaches, thus evaluating their feasibility of being entirely executed on mobile devices.
\end{itemize}

The remainder of the paper is organized as follows.
Section~\ref{sec:related} presents the related work.
In Section~\ref{sec:proposal}, we describe in details the procedure we follow to compare the different approaches for COVID-19 detection, including pre-processing of the raw audio sample, hand-crafted features extraction, and the use of deep audio embeddings models relying on the two transfer learning techniques.
Then, Section~\ref{sec:experiments} outlines the experimental setup we adopt to evaluate the classification performance of the considered approaches, and presents the obtained results.
In Section~\ref{sec:memory}, we evaluate the memory footprint of the two transfer learning techniques, considering their implementation on mobile devices.
Lastly, in Section~\ref{sec:conclusions}, we draw our conclusions and present some directions for future works.

\section{Related Work}
\label{sec:related}


It is largely recognized in literature that the use of deep learning techniques (namely, Convolutional Neural Networks, CNNs) for solving audio classification problems typically obtains better performances than classical ML methods~\cite{SSKP16}.
However, labelled datasets available for research on the application of ML to automated COVID-19 detection usually involve just a few hundreds of samples.
In such cases, \emph{transfer learning} is the main technique for system training so as to achieve good predictions in case of scarce labelled samples~\cite{pmlr-v27-bengio12a}.  
Leveraging transfer learning with CNNs for audio classification problems is studied in some papers.
In~\cite{tsalera2021comparison}, the usage of both image and sound CNNs is studied.
In the former case, data samples are image-based sound representations such as spectrograms.
Exhaustive experiments are conducted, which show that both hyper-parameters and the choice of the retrained part of the neural network impact on accuracy and retraining time, but in all the cases transfer learning yields better results than training a CNN from scratch.
The sound CNNs (namely VGGish and YAMNet~\cite{yamnet}) obtained better results than image CNNs (GoogLeNet, SqueezeNet, ShuffleNet) with all the considered datasets.

Transfer learning leverages neural networks, pre-trained on large sets (millions) of generic labelled objects, as a starting point.  Let $\cal{G}$ be the generic dataset used to pre-train the network $\cal{N}$, and $\cal{S}$ the dataset specific to the considered classification problem.  Two techniques are mainly used for transfer learning: either \emph{feature extraction} or \emph{fine-tuning}.
In the former, the first $n$ layers of $\cal{N}$ -- usually stopping prior the fully-connected layers -- receive as input the data in $\cal{S}$, and their outputs are used as features of the objects in $\cal{S}$. Thanks to the pre-training of $\cal{N}$, these should be the relevant features of $\cal{S}$ objects.  The remaining layers are then trained for the specific classification problem.  In the latter, the last (or a few last), most specialized, layer of $\cal{N}$ is removed and substituted with a new, non-trained, one; the resulting network $\cal{N’}$ is then trained on data from $\cal{S}$ (thus avoiding starting from scratch).

The superiority of one technique over the other is not definitively established.  Feature extraction is largely believed to avoid overfitting, i.e., excessive specialization of the network in properly classifying data with the same characteristics as those from the training set while poorly behaving with additional data of the same set; this phenomenon may easily occur with small datasets.  
In the computer vision field~\cite{ yosinski2014transferable}, the two methods are compared, and an analysis is conducted about where to split $\cal{N}$ for an effective feature extraction.  The drawn conclusions are that, in general, fine-tuning performs better than feature extraction; however, this benefit is smaller the more $\cal{G}$ and $\cal{S}$ differ.  Furthermore, the benefit depends on the number $n$ of preserved layers and it may drop for greater $n$ due to too few general features and/or complex feature interactions.
Similar hypotheses appear in~\cite{ZLLS22} (sec.14.2.1) where an assumption for fine-tuning usage is that the knowledge learned and preserved by $\cal{N’}$ is also applicable to $\cal{S}$.  
In the NLP field, the authors of~\cite{PRS19} obtain comparable performances of the two methods in most of the cases, with fine-tuning outperforming feature extraction when $\cal{G}$ and $\cal{S}$ are aligned, while achieving poor performance in case of significantly different tasks.
~\cite{9241777} analyzes the performances of fine-tuning applied to medical image processing.
The presented results confirm that layers selection in fine-tuning affects the classification metrics, and an automatic method for layer selection is proposed and validated, which consists in solving an optimization problem.
The proposed method shows an improvement for all the considered metrics, but at the cost of higher time spent for training.  An analogue research is presented in~\cite{GSKGRF19} applied to computer vision.

Going through the specific application domain of respiratory sounds, some solutions exist in the literature, even though it may be difficult to classify these sounds due to their complicated structures and variety of noises~\cite{ KHJLYCH21}.  
~\cite{CWYL22} classifies lung sounds via feature extraction applied to pre-trained CNNs.
It compares different techniques of feature extraction, considering three classifiers.
It obtains the best results with the pre-trained ResNet18 CNN and short-time Fourier transformation as the mechanism for feature extraction, and it improves the state-of-the-art results on the same dataset.
In~\cite{ KHJLYCH21}, deep learning pre-trained CNNs are used, and transfer learning is leveraged via feature extraction, representing sounds with Mel-spectrograms.
CNNs show better results than Support Vector Machine (SVM), and VGG-16 supplies the best results as feature extractor, with lower computation time than SVM; anyway, the best measured accuracy in detecting anomalous sounds is around $86\%$ confirming the difficulty of the considered classification task.
In~\cite{DSB20}, CNNs are used to classify lung diseases using audio classification, with a dataset of 920 samples reportedly challenging for other methods.
Comparison is conducted between two policies: $(i)$ performing feature extraction via a pre-trained (VGG-16) CNN whose output is then sent to SVM, or $(ii)$ fine-tuning the pre-trained CNN using spectrograms of lung sounds.
Both policies obtain better results than the state-of-the-art for the same dataset, with the former policy slightly outperforming (2.4$\%$) the latter. 

In the last couple of years,  researchers have explored several audio processing techniques to develop effective and low-cost COVID-19 screening methods based on respiratory data~\cite{DESHPANDE2022108289}, especially derived from smartphone embedded microphones.
One of the most promising approach in this domain consists in converting the audio files into a visual representation (e.g., time-frequency spectrogram or Mel-spectrogram) that can be used as input to a CNN model for both features extraction and classification.
This category includes, for example, the application AI4COVID proposed in~\cite{IMRAN2020100378} based only on cough recordings.
Specifically, they modelled the audio sample as both Mel-spectrogram and MFCC, which are then processed by an ensemble model composed of two CNNs and one SVM to categorize the cough into 4 classes: COVID-19, bronchitis, pertussis, and normal cough.
A similar solution has been proposed in~\cite{Mohammed2021}, where different visual representations of cough recordings (e.g., Mel-spectrogram, Chromagram, and Power-Spectrogram) have been compared to train an end-to-end CNN architecture.
Such approaches are particularly interesting because they avoid the features engineering and selection phases in the data processing pipeline, mainly relying on the intrinsic capabilities of DL to automatically model the raw input data.
However, due to the scarcity of public COVID-19 respiratory sound data, their training has been performed on small-size datasets, typically composed by a few hundreds of samples.
DL models, especially those with complex architectures, tend to overfit in such settings, often providing unreliable results.

By contrast, in~\cite{Han2022} a large, crowdsourced, dataset is available. Cough, breathing, and voice recordings of the participants are converted in log-mel spectrograms, which feed VGGish~\cite{7952132}, a CNN-based embedding model pre-trained on the large-scale YouTube-8M dataset (approximately 2.6 billion audio/video features), for feature extraction.
The obtained feature vectors are then aggregated by a pooling layer; the resulting latent feature vector is passed on to the prediction layers.
Thanks to the extension of the available dataset, no cross-validation is used.
Instead, participants whose data are included in the training set, are not considered for the testing set.
Exhaustive experiments are conducted; the results reveal that the considered dataset is challenging: a maximum ROC-AUC of $0.71$ is obtained in average, leveraging all the three data sources mentioned above.
By splitting the participants into categories, the best results are achieved for female participants and for over-60 participants.
The considered method is also validated by intentionally introducing biases into the data and repeating the experiments.
In Section~\ref{sec:experiments}, we present the results we obtained by conducting some experiments with the~\cite{Han2022} dataset.

Another approach deals with the mentioned DL drawback by using a combination of hand-crafted acoustic features and audio embeddings extracted by pre-trained deep models.
Representative of this category is~\cite{exploringcovid2020}, in which the authors used a set of acoustic features and a pre-trained DL model to train a shallow ML classifier (e.g, Logistic Regression, LR) to identify COVID-19 subjects from cough and breath audio recordings.
Specifically, as deep features extraction model, they employed VGGish, thus taking advantage of Transfer Learning concept to deal with the scarcity of COVID-19 audio data~\cite{9134370}.


\textcolor{blue}{Our work stands out from the related literature for the type of analysis we conduct in order to identify an effective solution to detect COVID-19 by using commercial mobile devices as low-cost and pervasive diagnostic tools. Indeed, while similar works consider just one or a few approaches of audio classification and evaluate their performance with a limited amount of data (typically only one dataset), here we provide a thorough comparison of several state-of-the-art techniques, including both shallow and deep learning models, by using four large-scale datasets collected from real devices, for a total of 13,448 audio samples.
Moreover, to the best of our knowledge, our work is the first investigating the two main approaches of Transfer Learning (i.e., features extraction and fine-tuning) in this specific application scenario, comparing their performance and providing a feasibility study of their implementation on real-world resource-constrained devices.}

\section{Acoustic features and Deep Audio Embeddings}
\label{sec:proposal}

\begin{figure}[t]
    \centering
    \includegraphics[width=\linewidth]{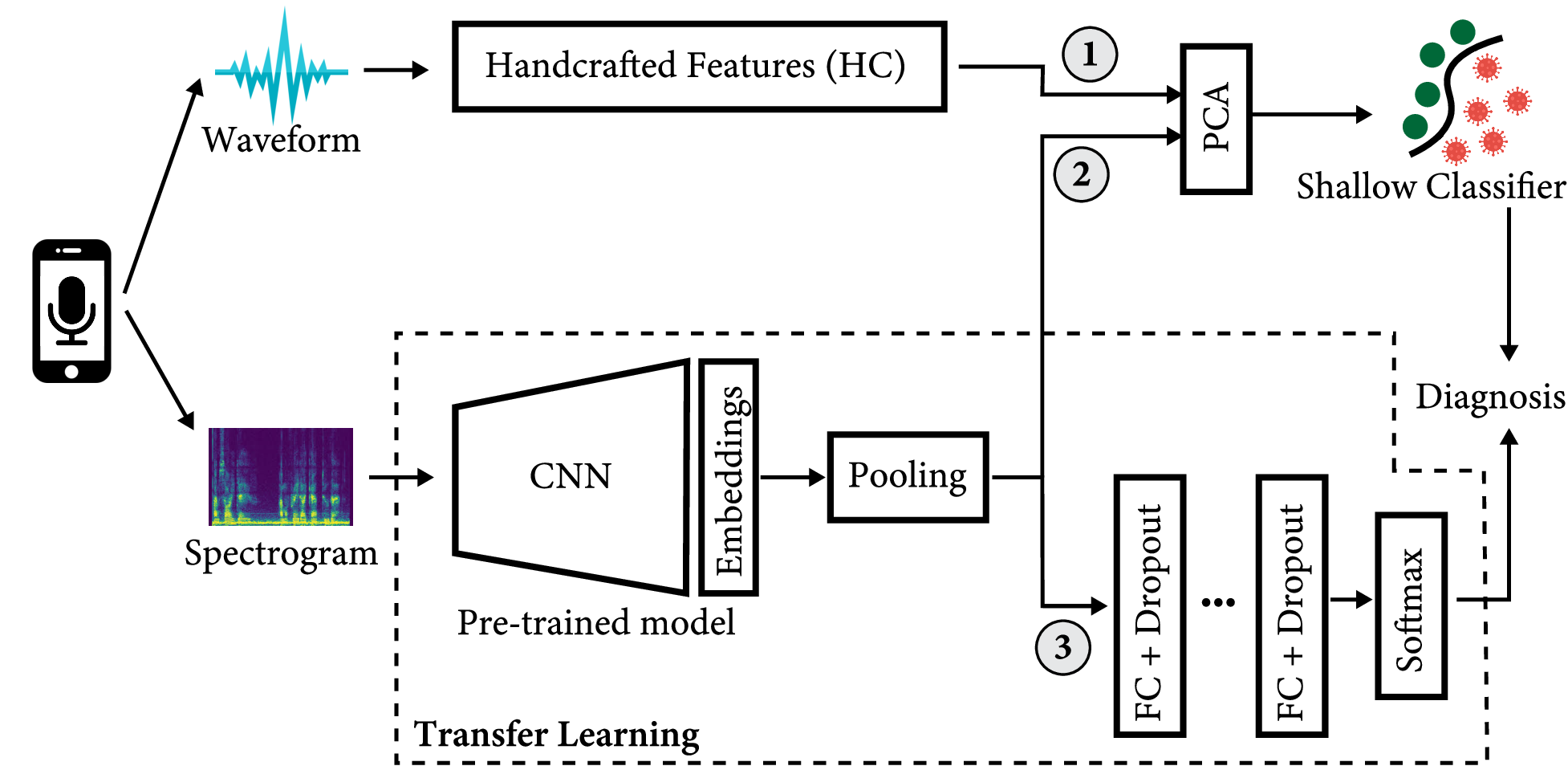}
    \caption{COVID-19 detection approaches: (1) extraction of handcrafted acoustic features from the audio waveform, which are then classified by a shallow ML model; (2) usage of a pre-trained deep learning model as features extractor, in series with a shallow model to classify the deep audio embeddings; and (3) fine-tuning of the pre-trained deep model for both features extraction and classification.}
    \label{fig:proposal}
\end{figure}



Figure~\ref{fig:proposal} shows the flow diagram of the entire data processing we adopt to compare the three main approaches of audio classification, which include: (1) using handcrafted acoustic features to train a shallow ML classifier; (2) using a pre-trained model as features extractor to input a shallow classifier; and, finally, (3) the fine-tuning of a pre-trained model, thus creating a single solution for both audio modelling and classification.

In the first approach, a set of handcrafted acoustic features is firstly extracted from the user's audio sample; then, a Principal Component Analysis (PCA) is applied to reduce the dimensionality of the features vector, which is finally used to input a shallow ML classifier that performs the diagnosis as a binary classification problem, distinguishing between COVID-positive and COVID-negative samples.

As far as transfer learning is concerned, after the signal conversion in Spectrogram-like images, and the extraction of deep audio embeddings, we use a pooling layer that calculates the mean and standard deviation of each dimension of the deep audio embeddings across all the windows. This generates a single feature vector. 

Then, based on the classification procedure, we can distinguish the two forms of transfer learning as follows: in (2), we first reduce the dimension of the embeddings vector through PCA to input a shallow ML classifier that provides the final prediction.
By contrast, in (3) we train from scratch a series of fully-connected layers to process the input embeddings vector, where the output softmax layer generates the final probability vector used to classify COVID-19 positive and negative predictions.

In the following, we describe in detail both the acoustic features and the pre-trained deep models considered in this work to evaluate the different approaches of audio classification aimed at detecting COVID-19 from respiratory sounds.

\subsection{Acoustic features extraction}



In order to transform the raw audio sample into a numerical representation manageable by a ML model, legacy approaches rely on different sets of handcrafted acoustic features.
To evaluate this approach, we adopt the common procedure used in similar audio-based medical applications~\cite{10.3389/fdgth.2020.00005}.
Firstly, the original audio sample recorded by the user's device microphone is re-sampled to a standard value for audio tasks (e.g., 16kHz or 22kHz).
Then, we manually extract common audio features related to both the frame (i.e., a chunk of the audio) and the segment (i.e., the entire audio sample) perspectives from the raw audio waveform, including frequency-based, structural, statistical, and temporal characteristics.

Specifically, we characterize the user's audio sample by using the same acoustic features proposed in~\cite{9414576}:

\begin{description}
\item[Duration]: total length (in seconds) of the audio sample;

\item[Onset]: number of pitch onsets (i.e., ``events'') in the audio file;

\item[Tempo]: the beats rate that occur at regular intervals throughout the whole audio signal;

\item[Period]: the frequency with the highest amplitude among those obtained from the Fast Fourier Transform (FTT);

\item[RMS Energy]: the root-mean-square value of the signal power, i.e., the magnitude of the short-time Fourier transform;

\item[Spectral Centroid]: the centroid value of the frame-wise magnitude spectrogram, which can be used to identify percussive and sustained sounds;

\item[Roll-off Frequency]: the frequency under which the 85\% of the total energy of the frame-wise spectrum is contained;

\item[Zero-crossing rate]: the number of times the signal value crosses the zero axes, and it is computed for each frame;

\item[Mel-Frequency Cepstral Coefficients (MFCC)]: coefficients that represent the shape of the cosine transformation of the sound logarithmic spectrum, expressed in Mel-bands, which characterize the set of frequency audible by human beings;

\item[$\Delta$-MFCC, $\Delta^2$-MFCC]: the first and second order derivatives of MFCC along the time dimension, respectively.

\end{description}

The total number of acoustic features we extract from the audio sample is 477, including standard statistics (e.g., mean, median, max/min values, and skewness) to describe time-series descriptors for the entire audio signal, i.e., for RMS Energy, Spectral Centroid, Roll-Off Frequency, Zero-crossing rate crossing, MFCC, $\Delta$-MFCC, and $\Delta^2$-MFCC.

\subsection{Pre-trained deep audio embeddings models}

In order to evaluate the effectiveness of the transfer learning approaches in our specific application scenario, we take into account 3 state-of-the-art deep learning models for audio classification: YAMNET, VGGish, and L\textsuperscript{3}-Net~\cite{jsan10040072}.
These models mainly differ in the network's architecture and training approach.

Specifically, VGGish~\cite{7952132} is a 24-layer-deep network based on the VGG architecture, a popular CNN-based model for image classification~\cite{simonyan2014very}.
The input audio sample must be firstly re-sampled at 16 kHz and normalized in the range [-1, +1].
Then, the audio waveform is framed into sliding windows of 0.96 seconds with no overlap.
Finally, for each frame, the corresponding Mel-spectrogram image is used to input VGGish, which relies on the CNN layers to produce a 128-dimensional embeddings representation of the input image, which can be further classified either by a set of fully-connected layers or a shallow ML model.

YAMNET follows the same idea of VGGish, but it is composed by 28 deep layers and it is based on the architecture of MobileNet~\cite{howard2017mobilenets}, a lightweight CNN model especially designed to perform visual applications directly on mobile devices.
In this case, the input waveform should be framed through a sliding windows of length 0.96 seconds and hop of 0.48 seconds, from which the model extracts a 1024-dimensional embeddings vector.

By contrast, L\textsuperscript{3}-Net is based on a totally different approach.
In fact, as we mentioned in Section~\ref{sec:related}, this model has been designed to learn deep representations by identifying if a video frame (i.e., an image) and an audio segment come from the same video source.
This allows to train the model by exploiting the self-supervised approach: since both matched and mismatched image-audio pairs can be automatically generated by extracting the image and audio from the same or different videos, no manual labeling is required to train the model.

The architecture of L\textsuperscript{3}-Net consists of two distinct CNN subnetworks, used to extract different embeddings for the video and audio inputs respectively.
To check the correspondence between both embeddings, a fusion network is used, which concatenates the deep representations and uses two fully connected layers as well as a softmax layer for binary classification.
After training, the two subnetworks can be used as two distinct models to extract deep audio and video embeddings, respectively.

L\textsuperscript{3}-Net has been further extended in~\cite{8682475}, where the authors investigated different design choices of the original model, and how they impact on the performances of downstream audio classifiers.
In particular, they studied 3 main aspects of the model: (i) the input representation, considering Linear-frequency power spectrogram (\texttt{L}), Mel-spectrogram with 128 bands (\texttt{M128}), and Mel-spectrogram with 256 bands (\texttt{M256}); (ii) the type of audio samples used during the training, including both environmental sounds (\texttt{E}) and music (\texttt{M}); and (iii) the size of the deep audio embeddings, considering both \texttt{512} and \texttt{6144} dimensions.
Based on this study, the authors publicly released OpenL3~\footnote{\url{https://openl3.readthedocs.io}}, a framework that includes all the 12 investigated configurations of L\textsuperscript{3}-Net, which we summarized in Table~\ref{tab:l3_conf} for the sake of clarity.

\begin{table}[t]
\caption{L\textsuperscript{3}-Net configurations.}
\label{tab:l3_conf}
\resizebox{\textwidth}{!}{%
\begin{tabular}{lccl}
\toprule
\textbf{Name}  & \textbf{Training data}                & \textbf{Embedding size} & \textbf{Input type} \\
\midrule
L3 E 512 L     & \multirow{6}{*}{\shortstack{Environmental \\ sounds}} & \multirow{3}{*}{512}    & Linear                    \\
L3 E 512 M128  &                                       &                         & Mel-128                   \\
L3 E 512 M256  &                                       &                         & Mel-256                   \\
\cmidrule{3-4}
L3 E 6144 L    &                                       & \multirow{3}{*}{6144}   & Linear                    \\
L3 E 6144 M128 &                                       &                         & Mel-128                   \\
L3 E 6144 M256 &                                       &                         & Mel-256                   \\
\cmidrule{1-4}
L3 M 512 L     & \multirow{6}{*}{Music}                & \multirow{3}{*}{512}    & Linear                    \\
L3 M 512 M128  &                                       &                         & Mel-128                   \\
L3 M 512 M256  &                                       &                         & Mel-256                   \\
\cmidrule{3-4}
L3 M 6144 L    &                                       & \multirow{3}{*}{6144}   & Linear                    \\
L3 M 6144 M128 &                                       &                         & Mel-128                   \\
L3 M 6144 M256 &                                       &                         & Mel-256                 \\
\bottomrule
\end{tabular}
}
\end{table}

In order to compare the different approaches of audio processing for COVID-19 detection, we rely on the pre-trained instances of the aforementioned deep learning models, thus taking advantage of their ability of characterizing audio data samples obtained by their training  with a massive amount of data.
In particular, for VGGish and YAMNET, we use the pre-trained versions available on \emph{TensforFlow Hub},\footnote{\url{https://www.tensorflow.org/hub}} which have been trained by using the YouTube-8M~\cite{abu2016youtube} (6.1 million videos) and AudioSet~\cite{10.3389/fdgth.2020.00005} (more than 2 million of data samples), respectively. We also test all the configurations of OpenL3, which have been trained on different subsets of AudioSet.

\section{Experimental evaluation}
\label{sec:experiments}

In this section, we describe both the benchmark datasets and the experimental protocol we use to evaluate the performances of the different approaches presented in Section~\ref{sec:proposal}.
In a first set of experiments, we compare the effectiveness of both handcrafted acoustic features and deep audio embeddings in detecting COVID-19 from respiratory sounds data.
In a second set of experiments, we evaluate the feasibility of fine-tuning the considered deep learning models with the available data, thus creating a single component for both features extraction and classification.

\subsection{Datasets}

In order to perform our experiments, we rely on 4 crowdsourced datasets commonly used in the literature to benchmark AI-based solutions for COVID-19 detection from respiratory sounds, including both breathing and coughing recordings: \texttt{COSWARA}~\cite{coswarads} and \texttt{Coughvid}~\cite{Orlandic2021} are publicly available, while we obtained the access to two different versions of the COVID-19 Sounds dataset collected by the Cambridge University,~\footnote{\url{https://www.covid-19-sounds.org}} here called \texttt{Cambridge KDD}~\cite{exploringcovid2020} and \texttt{Cambridge npj}~\cite{Han2022}.

The first dataset is part of the \texttt{COSWARA} research project of the Indian Institute of Science (IISc), Bangalore, attempting to build a diagnostic tool for COVID-19 using different audio recordings of individuals, including breathing, cough and speech sounds.
Currently, the project is still ongoing and it is continuing the data collection stage through crowdsourcing.
Through the use of a web and a mobile application, the researchers ask volunteers to send their health status along with different types of audio recordings: two samples of cough (shallow and heavy), two audios of breath (shallow and deep), two recordings of counting numbers (normal and fast), and the phonation of sustained vowels.

The dataset is freely available on the official Github repository of the project~\footnote{\url{https://github.com/iiscleap/Coswara-Data}}.
We downloaded the whole dataset available on February, 26th 2022, which includes audio recordings collected up to 24th February 2022.
In this work, we considered only valid files (i.e., not corrupted or empty recordings) of both breath and cough recordings donated by people without other respiratory diseases (e.g., asthma), including both the available modalities, i.e., shallow/heavy cough, and shallow/deep breath.
The final dataset includes a total of 2501 audio samples shared by people who have declared they were positive to COVID-19 at record time (1234 breath samples, and 1267 cough samples), and 860 samples labeled as negative examples (425 breath and 435 cough).

Similarly to \texttt{COSWARA}, the \texttt{Coughvid} corpus has been collected through a web application by a research group of the École Polytechnique Fédérale de Lausanne, Switzerland, between April 1st, 2020 and December 1st, 2020, and it is freely available as a Zenodo repository~\footnote{\url{https://doi.org/10.5281/zenodo.4048312}}.
The authors collected a total of 25,000 crowdsourced samples representing a wide range of participant ages, genders, and geographic locations.
Moreover, among the different metadata, each record has been labeled according to the output of a ML classifier specifically trained by the authors to estimate the probability that the input audio sample actually represents a cough recordings.
Taking into account only those records with a high cough estimation (i.e., greater than 80\%), we selected a total of 547 COVID-positive samples and 5,625 negative recordings.

Finally, both \texttt{Cambridge KDD} and \texttt{Cambridge npj} have been collected by the Mobile System Research Lab of the University of Cambridge as part of the ERC EAR research project.
Similarly to the previous datasets, they contain respiratory sounds collected by using both web and mobile applications.
\texttt{Cambridge KDD} represents the first dataset used by the authors to evaluate the performance of a VGGish-based solution to detect COVID-19 from both breathing and coughing recordings~\cite{exploringcovid2020}, and it contains 282 COVID-positive samples (141 breath and 141 cough) and 660 negative samples (330 breath and 330 cough).
More recently, in~\cite{Han2022} the same authors have studied how different data selection strategies can overestimate the performances of ML models in detecting COVID-19 from respiratory sounds, and they released a second dataset to avoid potential acoustic bias, selecting only English-speaking participants and balancing the data based on the users' demographics~\footnote{While the access to the raw data requires a formal Data Access Agreement, the code required to produce the balanced dataset is freely available on Github at~\url{https://github.com/cam-mobsys/covid19-sounds-npjDM}}.
In this work, we refer to such corpora as \texttt{Cambridge npj} to distinguish it from the former dataset, and it contains a total of 1468 (734 breath and 734 cough) audio samples donated by COVID-positive subjects and 1504 (752 breath and 752 cough) recordings of healthy people.

\begin{figure}[t]
    \centering
    \includegraphics[width=0.95\linewidth]{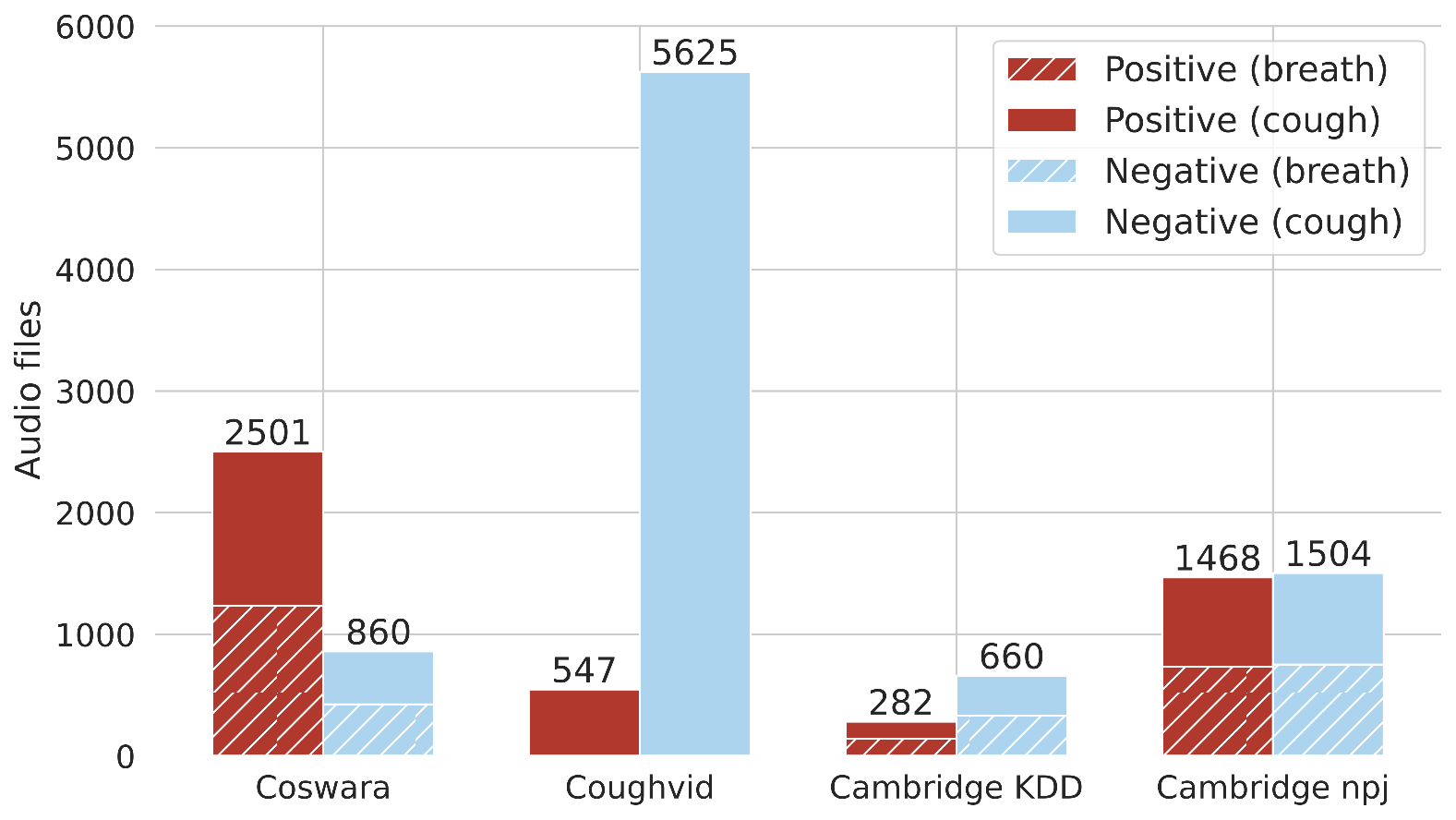}
    \caption{Number of audio samples in the 4 considered datasets, grouped by their respective labels and modalities.}
    \label{fig:datasets}
\end{figure}

Figure~\ref{fig:datasets} shows the main peculiarities of the four datasets, highlighting the number of audio files available for both negative and positive examples, according to the available ground truth.
As we can note, most of the datasets are characterized by a large class imbalance, which can lead to obtain biased learning models that have a poorer predictive accuracy over the minority classes compared to the majority classes~\cite{Zheng2020}.
To cope with this issue, we balanced all the datasets by using random under sampling, thus randomly removing samples from the majority class in each corpus.
In order to make the experiments reproducible, we share both the name of the files used in our experiments and the code on our Github repository~\footnote{\url{https://github.com/mattiacampana/Transfer-Learning-COVID-19}}.

\subsection{Evaluation protocol}
\label{sec:eval_protocol}

We follow a rigorous protocol for the experiment's evaluation.
\textcolor{blue}{Specifically, while similar works in the literature employ simple dataset splits that can lead to overestimate the performance of a predictor, we decided to rely on the 5-fold Nested Cross-Validation approach with user-based splits to avoid biasing the results based on patterns of specific users or specific train/validation/test splits of the dataset. The procedure can be summarized as follows.}
Firstly, we divide the whole dataset into 5 non-overlapping folds, ensuring that all the recordings generated by the same user are not placed in two different folds.
In an outer loop, we iterate over the 5 folds, where, for each iteration, one fold is selected as the test set, while the remaining 4 represent the development set.
The development set is then used in an inner loop for hyperparameters tuning, where it is further divided in 5 non-overlapping folds to define both training and validation sets.

The second step of the evaluation protocol differs based on the Transfer Learning approach.
For the features extraction experiments, where the pre-trained deep embeddings models are simply used as feature extractors, the tuning phase includes: (i) finding the best PCA coefficient, that is, the amount of variance that needs to be explained by the held components; and (ii) finding the best shallow classifier and tuning its parameters by using the \emph{Bayesian optimization algorithm}~\cite{NIPS2012_05311655}.
In these experiments, we test 4 broadly used ML algorithms: Logistic Regression (LR), Support Vector Machines (SVMs), Random Forest (RF), and AdaBoost (AB).

\begin{table}[t]
\caption{Hyperparameters tuned in both the Features Extraction (FE) and Fine-Tuning (FT) experiments. Algorithm DE indicates the classification layers added to the Deep Embeddings models considered in our experiments.}
\label{tab:grid_search}
\centering
\resizebox{\textwidth}{!}{%
\begin{tabular}{llll}
\toprule
Task & Algorithm            & Parameter           & Values                            \\
\midrule
\multirow{13}{*}{FE}
& \multirow{4}{*}{SVM}
                        & regularization            & {[}$10^{-3}, \dotsc, 10^3${]}         \\
&                       & kernel                    & {[}rbf, poly, sigmoid{]}              \\
&                       & kernel coefficient        & {[}$10^{-3}, \dotsc, 10${]}           \\
&                       & degree of poly kernel     & {[}$2, \dotsc, 5${]}                  \\
\cmidrule{2-4}
& \multirow{2}{*}{AB}
                        & estimators                & {[}10, 20, 50, 100{]}                 \\
&                       & learning rate             & {[}1, .5, .1, .05, .01, .001{]}       \\
\cmidrule{2-4}
& \multirow{2}{*}{LR}
                        & penalty                   & {[}l1, l2{]}                      \\
&                       & regularization            & {[}$10^{-3}, \dotsc, 10^3${]}     \\
\cmidrule{2-4}
& \multirow{4}{*}{RF}
                        & estimators                & {[}10, 20, 50, 100{]}             \\
&                       & min samples split         & {[}2, 8, 10, 12{]}                \\
&                       & max depth                 & {[}10, 30, 50{]}                  \\
&                       & split criterion           & {[}entropy, gini{]}           \\
\cmidrule{2-4}
& PCA                  & explained variance           & {[}.6, .65, .7, .75, .8, .85, .9. .95, .99{]}        \\
\midrule
\multirow{3}{*}{FT}
& \multirow{3}{*}{DE}
                     & hidden layers                & {[}$1, \dotsc, 5${]}                  \\
&                    & hidden units                 & {[}128, 512, 1024, 2048, 6144{]}      \\
&                    & dropout rate                 & {[}0, .1, .2, .3, .4{]}      \\
\bottomrule
\end{tabular}
}
\end{table}

By contrast, in the set of fine-tuning experiments, we tune the considered deep embedding models by using the \emph{Hyperband} algorithm, which is broadly used to optimize neural network models because it is able to speed up the random search of the parameter spaces through adaptive resource allocation and early-stopping~\cite{li2017hyperband}.
Specifically, we exploit such optimization algorithm to tune the hyperparameters of the additional classification layers added to the base deep embeddings models, including the number of fully-connected layers, the number of hidden units in each layer, and the percentage of dropout rates to control the over-fitting of the model.

Table~\ref{tab:grid_search} summarizes the value spaces of the hyperparameters considered during the tuning phase of both features extraction and fine-tuning experimental settings.
In addition, for each model, we also evaluate which type of audio files (i.e., modality) allows us to obtain the best performances among those available in each dataset: \emph{Cough} (C), \emph{Breath} (B), or the combination of the two (CB).

Finally, we evaluate the classification performances over the outer splits by using the Precision-Recall Area Under Curve (PR-AUC), which summarizes the Precision-Recall curve with a wide range of threshold values as a single score, and it is commonly used to evaluate binary classification models with imbalanced datasets~\cite{10.1145/2907070}.

\subsection{Features extraction results}

\begin{figure}[t]
    \centering
    \includegraphics[width=\linewidth]{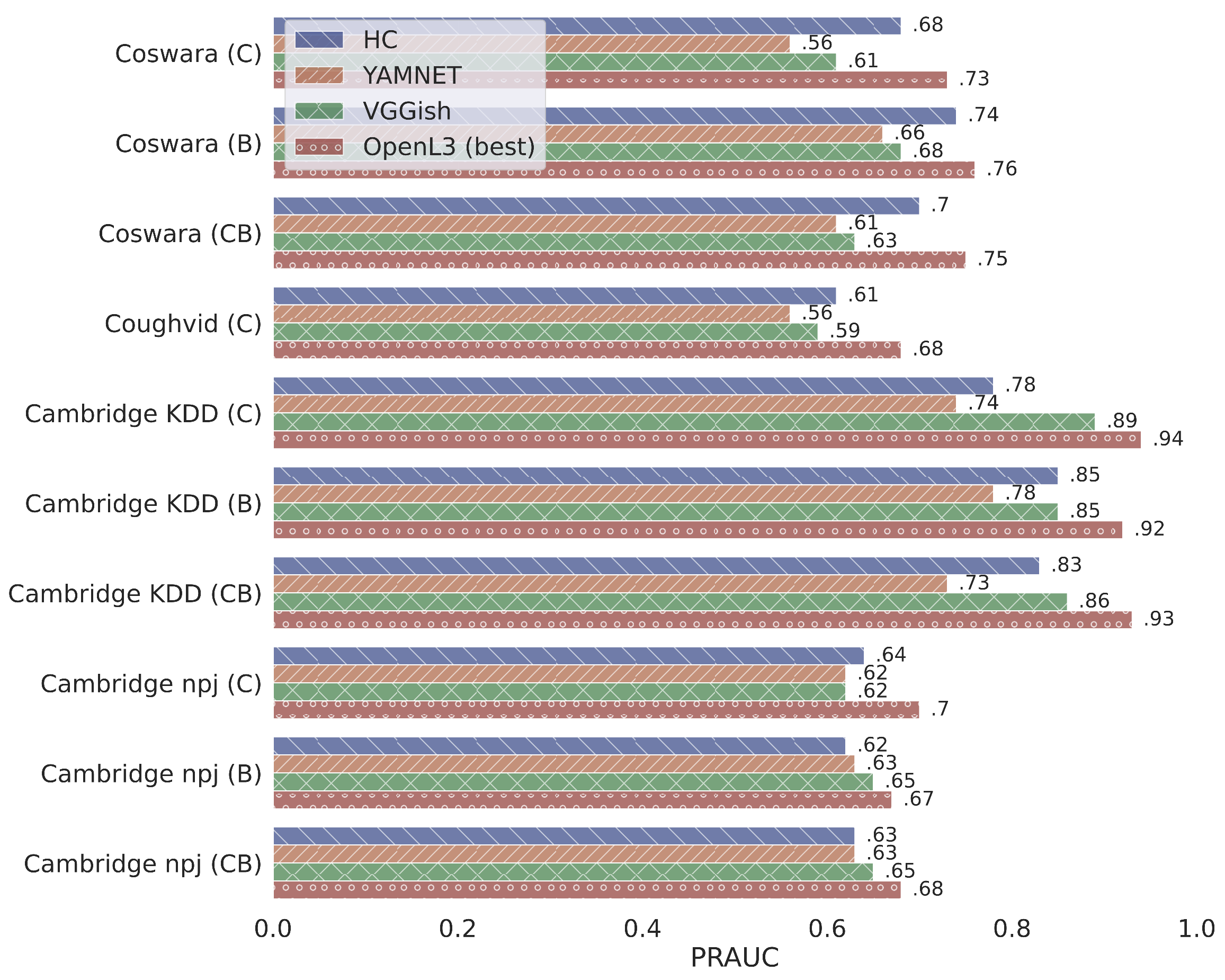}
    \caption{Overall results for the Features Extraction experiments. OpenL3 (best) refers to the model's configuration that has obtained the best result in each dataset/modality.}
    \label{fig:features_extraction_overall_results}
\end{figure}

In the first set of experiments we compare the quality of different audio features as input of standard shallow classifiers in COVID-19 detection task.
Such features include both the hand-crafted acoustic features (HC) presented before, and the audio embeddings extracted by YAMNET, VGGish, and OpenL3.

Figure~\ref{fig:features_extraction_overall_results} summarizes the obtained results with all the considered datasets and the relative audio modalities, indicating as \emph{OpenL3 (best)} the configuration of OpenL3 that presents the best performances.
As we can note, the performances of each approach strongly differs for each combination of dataset and audio modality but, in general, OpenL3 is able to obtain the best results in all the experimental settings.
Compared with the other deep models, the embeddings of OpenL3 allows to achieve an overall improvement of 12.3\% with respect to the deep features generated by YAMNET, and of 7.2\% more than those extracted by VGGish.
Surprisingly, the lowest improvement obtained by OpenL3 is the one compared with HC (6.7\%), which perform better than YAMNET and VGGish in all the experimental settings, except for \texttt{Cambridge npj (B)} and \texttt{Cambridge npj (CB)}, where the handcrafted features achieve the same or slightly worse results than the two deep embedding models.
According to the obtained results, \texttt{Cambridge npj} represents a challenging dataset, where all the considered solutions achieve lower performances compared with the results achieved in the other experimental settings, ranging from 0.62 and 0.67 of PR-AUC, except for OpenL3 that can reach 0.70 by using only cough samples.

\begin{table}[t]
\caption{Features extraction results with Coswara and Coughvid.}
\label{tab:fe_coswara_coughvid}
\resizebox{\textwidth}{!}{%

\begin{tabular}{ll|crr|crr|crr}
\toprule
& & \multicolumn{3}{c|}{\textbf{Cough}} & \multicolumn{3}{c|}{\textbf{Breath}} & \multicolumn{3}{c}{\textbf{Both}}                     \\
\textbf{D} & \textbf{Features} & \textbf{Clf} & \textbf{PCA} & \textbf{PR-AUC} & \textbf{Clf} & \textbf{PCA} & \textbf{PR-AUC} & \textbf{Clf} & \textbf{PCA} & \textbf{PR-AUC} \\
\toprule
\multirow{15}{*}{\rotatebox[origin=c]{90}{Coswara}}
& HC                & AB           & .90          & .68            & AB           & .99          & .74            & SVM          & .65          & .70            \\
& YAMNET            & RF           & .99          & .56            & LR           & .99          & .66            & RF           & .99          & .61            \\
& VGGish            & RF           & .70          & .61            & RF           & .75          & .68            & SVM          & .65          & .63            \\
\cmidrule{2-11}
& L3 E 512 L        & \textbf{LR}  & \textbf{.70} & \textbf{.73}   & RF           & .90          & .70            & AB           & .99          & .74   \\
& L3 E 512 M128     & LR           & .99          & .71            & RF           & .75          & .70            & AB           & .95          & .73            \\
& L3 E 512 M256     & SVM          & .85          & .68            & AB           & .95          & .71            & AB           & .85          & .72            \\
& L3 E 6144 L       & LR           & .95          & .65            & LR           & .99          & .69            & LR           & .99          & .68            \\
& L3 E 6144 M128    & LR           & .85          & .72            & \textbf{AB}  & \textbf{.99} & \textbf{.76}   & SVM          & .75          & .73 \\
& L3 E 6144 M256    & LR           & .75          & .72            & RF           & .85          & .74            & AB           & .99          & .71            \\
& L3 M 512 L        & RF           & .70          & .71            & RF           & .85          & .73            & SVM          & .85          & .70            \\
& L3 M 512 M128     & RF           & .80          & .69            & AB           & .99          & .74            & RF           & .70          & .72            \\
& L3 M 512 M256     & LR           & .60          & .70            & RF           & .90          & .71            & AB           & .99          & .69            \\
& L3 M 6144 L       & LR           & .75          & .70            & LR           & .99          & .70            & AB           & .95          & .72            \\
& L3 M 6144 M128    & SVM          & .75          & .69            & LR           & .99          & .74            & AB           & .99          & .72            \\
& L3 M 6144 M256    & SVM          & .99          & .69            & AB           & .90          & .72            & \textbf{RF}  & \textbf{.85} & \textbf{.75}   \\
\midrule

\multirow{15}{*}{\rotatebox[origin=c]{90}{Coughvid}}
& HC                & SVM          & .80          & .61            & -            & -            & -              & -            & -            & -             \\
& YAMNET            & SVM          & .60          & .56            & -            & -            & -              & -            & -            & -             \\
& VGGish            & LR           & .70          & .59            & -            & -            & -              & -            & -            & -             \\
\cmidrule{2-11}
& L3 E 512 L        & SVM          & .75          & .62            & -            & -            & -              & -            & -            & -             \\
& L3 E 512 M128     & LR           & .99          & .67            & -            & -            & -              & -            & -            & -             \\
& L3 E 512 M256     & LR           & .99          & .64            & -            & -            & -              & -            & -            & -             \\
& L3 E 6144 L       & LR           & .99          & .65            & -            & -            & -              & -            & -            & -             \\
& L3 E 6144 M128    & \textbf{LR}  & \textbf{.99} & \textbf{.68}   & -            & -            & -              & -            & -            & -             \\
& L3 E 6144 M256    & LR           & .99          & .67            & -            & -            & -              & -            & -            & -             \\
& L3 M 512 L        & RF           & .90          & .62            & -            & -            & -              & -            & -            & -             \\
& L3 M 512 M128     & LR           & .99          & .62            & -            & -            & -              & -            & -            & -             \\
& L3 M 512 M256     & LR           & .75          & .60            & -            & -            & -              & -            & -            & -             \\
& L3 M 6144 L       & LR           & .99          & .67            & -            & -            & -              & -            & -            & -             \\
& L3 M 6144 M128    & LR           & .99          & .65            & -            & -            & -              & -            & -            & -             \\
& L3 M 6144 M256    & LR           & .99          & .64            & -            & -            & -              & -            & -            & -             \\
\bottomrule
\end{tabular}
}
\end{table}

In order to analyze in detail the different solutions, we also report the shallow classifiers and PCA explained variance rate that led to the best results.
Specifically, Table~\ref{tab:fe_coswara_coughvid} shows the details for the \texttt{Coswara} and \texttt{Coughvid} datasets, while Table~\ref{tab:fe_cambridge} shows those with \texttt{Cambridge KDD} and \texttt{Cambridge npj}.
With the former dataset, the best configuration is different for each modality.
HC performs best with Breath recordings, reaching a PR-AUC score of 0.74 by using all the features (i.e., 99\% PCA) and AB as shallow classifier, while it obtains its lowest score (0.68) using the Cough modality.
Also YAMNET obtains its best result with the breathing recordings, but showing a significant decrease in terms of PR-AUC, obtaining an overall score of 0.66 with LR.
VGGish performs slighlty better, reaching 0.68 of PR-AUC using RF and 75\% as PCA explained variance.
On the other hand, OpenL3 obtains the best performances with all the three modalities, but with different configurations: using only cough samples, the best configuration is \texttt{L3 E 512 L}, which reaches 0.73 PR-AUC by using LR and 0.70 of PCA; \texttt{L3 E 6144 M128} reaches the best score over the whole dataset (i.e., 0.76 PR-AUC) relying on breathing sounds and AB as classifier; while \texttt{L3 M 6144 M256} scores 0.75 of PR-AUC by using RF and 85\% of PCA with the combination of the two previous modalities.

In the experiments with the \texttt{Coughvid} dataset, all the solutions are characterized by a slight decrease in the performance: YAMNET scores 0.56 of PR-AUC by using SVM, thus representing the worst model; VGGish performs better, reaching 0.59 with LR; HC still obtains better performance compared with the previous two deep models, reaching a PR-AUC of 0.61 with SVM; while OpenL3 overcomes the other solutions in all the tested configurations, starting from a PR-AUC of 0.62 obtained by \texttt{L3 E 512 L} in combination with SVM, up to the best result of 0.68 PR-AUC scored by \texttt{L3 E 6144 M128} with the simple LR as shallow classifier.

\begin{table}[t]
\caption{Features extraction results with Cambridge KDD and Cambridge npj.}
\label{tab:fe_cambridge}
\resizebox{\textwidth}{!}{%

\begin{tabular}{ll|crr|crr|crr}
\toprule
& & \multicolumn{3}{c|}{\textbf{Cough}} & \multicolumn{3}{c|}{\textbf{Breath}} & \multicolumn{3}{c}{\textbf{Both}}                     \\
\textbf{D} & \textbf{Features} & \textbf{Clf} & \textbf{PCA} & \textbf{PR-AUC} & \textbf{Clf} & \textbf{PCA} & \textbf{PR-AUC} & \textbf{Clf} & \textbf{PCA} & \textbf{PR-AUC} \\
\toprule

\multirow{15}{*}{\rotatebox[origin=c]{90}{Cambridge KDD}}
& HC                & AB           & .60          & .78            & AB           & .80          & .85             & RF           & .60          & .83   \\
& YAMNET            & LR           & .99          & .74            & SVM          & .99          & .78             & LR           & .95          & .73   \\
& VGGish            & AB           & .99          & .89            & LR           & .65          & .85             & AB           & .95          & .86   \\
\cmidrule{2-11}
& L3 E 512 L        & LR           & .70          & .86            & LR           & .95          & .87             & RF           & .90          & .88   \\
& L3 E 512 M128     & LR           & .60          & .85            & LR           & .85          & .88             & LR           & .80          & .86   \\
& L3 E 512 M256     & RF           & .80          & .91            & SVM          & .85          & .88             & AB           & .99          & .89   \\
& L3 E 6144 L       & RF           & .75          & .88            & RF           & .75          & .84             & SVM          & .99          & .86   \\
& L3 E 6144 M128    & RF           & .70          & .77            & AB           & .85          & .83             & AB           & .90          & .81   \\
& L3 E 6144 M256    & \textbf{LR}  & \textbf{.90} & \textbf{.94}   & \textbf{LR}  & \textbf{.80} & \textbf{.92}    & \textbf{LR}  & \textbf{.85} & \textbf{.93}   \\
& L3 M 512 L        & LR           & .70          & .82            & RF           & .80          & .89             & SVM          & .65          & .86   \\
& L3 M 512 M128     & SVM          & .95          & .77            & SVM          & .60          & .81             & SVM          & .99          & .80            \\
& L3 M 512 M256     & LR           & .95          & .86            & RF           & .95          & .87             & LR           & .90          & .89   \\
& L3 M 6144 L       & LR           & .75          & .91            & SVM          & .85          & .89             & LR           & .70          & .88   \\
& L3 M 6144 M128    & RF           & .90          & .93            & \textbf{LR}  & \textbf{.85} & \textbf{.92}    & AB           & .95          & .91   \\
& L3 M 6144 M256    & LR           & .65          & .92            & \textbf{LR}  & \textbf{.90} & \textbf{.92}    & \textbf{LR}  & \textbf{.85} & \textbf{.93}   \\

\midrule

\multirow{15}{*}{\rotatebox[origin=c]{90}{Cambridge npj}}
& HC                 & SVM          & .70          & .64           & SVM          & .99          & .62              & LR          & .70          & .63   \\
& YAMNET             & SVM          & .60          & .62           & SVM          & .99          & .63              & SVM         & .75          & .63   \\
& VGGish             & SVM          & .70          & .62           & SVM          & .99          & .65              & SVM         & .65          & .65   \\
\cmidrule{2-11}
& L3 E 512 L         & \textbf{SVM} & \textbf{.75} & \textbf{.70}  & SVM          & .65          & .65              & SVM         & .65          & .63   \\
& L3 E 512 M128      & SVM          & .99          & .67           & LR           & .99          & .64              & SVM         & .90          & .67   \\
& L3 E 512 M256      & LR           & .99          & .65           &\textbf{SVM}  & \textbf{.99} & \textbf{.67}     & SVM         & .90          & .65   \\
& L3 E 6144 L        & LR           & .95          & .65           & SVM          & .80          & .63              & LR          & .80          & .65   \\
& L3 E 6144 M128     & LR           & .95          & .65           & SVM          & .75          & .62              & SVM         & .85          & .66   \\
& L3 E 6144 M256     & LR           & .95          & .66           & LR           & .85          & .64              & SVM         & .80          & .66   \\
& L3 M 512 L         & SVM          & .95          & .65           & SVM          & .65          & .62              & SVM         & .90          & .64   \\
& L3 M 512 M128      & LR           & .99          & .66           & LR           & .60          & .62              & SVM         & .80          & .64   \\
& L3 M 512 M256      & SVM          & .65          & .66           & SVM          & .90          & .66              & SVM         & .75          & .64   \\
& L3 M 6144 L        & SVM          & .90          & .65           & SVM          & .65          & .62              & LR          & .99          & .64   \\
& L3 M 6144 M128     & LR           & .95          & .68           & SVM          & .85          & .66              &\textbf{SVM} &\textbf{.85}  &\textbf{.68}   \\
& L3 M 6144 M256     & LR           & .95          & .67           & SVM          & .90          & .63              & SVM         & .75          & .66   \\
\bottomrule
\end{tabular}
}
\end{table}

On the other hand, with \texttt{Cambridge KDD} all the considered features models reach higher performances with all the three modalities.
The lowest value scored by HC is 0.78 by using AB as classifier and coughing records only; while it reaches its maximum result with the breath modality, obtaining 0.85 of PR-AUC and AB.
YAMNET still represents the worst model, obtaining the best results with breath sounds: 0.78 of PR-AUC reached by using SVM as shallow classifier.
In this case, VGGish performs better than HC and YAMNET, reaching a PR-AUC of 0.89 with cough recordings and the classifier AB.
OpenL3 still greatly overcomes the other solutions, obtaining its lowest score with the Breath modality (i.e., 0.92 of PR-AUC) with LR as classifier and three different configurations (i.e., \texttt{L3 E 6144 M256}, \texttt{L3 M 6144 M128}, and \texttt{L3 M 6144 M256}); while it reaches the best performance with the \texttt{L3 E 6144 M256} configuration and LR as classifier, obtaining an overall PR-AUC of 0.94.

Finally, in the \texttt{Cambridge npj} experiments, HC, YAMNET, and VGGish obtain very similar results with all the three modalities, while OpenL3 still represents the best model.
Specifically, HC shows the best performances with coughing samples only, scoring 0.64 of PR-AUC with SVM and 70\% of PCA.
YAMNET performs very similarly, reaching 0.63 as PR-AUC as maximum value, but using only breathing samples and the combination of cough and breath modalities.
With the same combination of respiratory sounds, VGGish slightly overcomes the previous two solutions, scoring a PR-AUC of 0.65 by using SVM.
Also in this case, OpenL3 generates the best features set, which allows to reach a PR-AUC score of 0.70 by using the \texttt{L3 E 512 L} configuration, the SVM classifier, and cough recordings only.

\subsection{Fine-tuning results}

Instead of using the considered deep audio embedding models as simple features extractors, in this second set of experiments we evaluate the use of transfer learning via fine-tuning to create a single model to both represent the input sound through deep embeddings, and then classify it.
In other words, here we start with the base models (YAMNET, VGGish, and OpenL3), pre-trained for generic audio classification tasks, and then we fine-tune (train) only the final fully-connected layers of the neural networks to detect COVID-19 from respiratory sounds data.

\begin{table}[t]
\caption{PR-AUC scores obtained in the fine-tuning experiments.}
\label{tab:fine_tuning_results}
\resizebox{\textwidth}{!}{%
\begin{tabular}{l|rrr|r|rrr|rrr}
\toprule
                  & \multicolumn{3}{c|}{\textbf{Coswara}}       & \textbf{Coughvid} & \multicolumn{3}{c|}{\textbf{Cambridge KDD}} & \multicolumn{3}{l}{\textbf{Cambridge npj}} \\
\textbf{Features} & \textbf{C}   & \textbf{B}   & \textbf{CB} & \textbf{C}                            & \textbf{C}   & \textbf{B}   & \textbf{CB} & \textbf{C}   & \textbf{B}   & \textbf{CB} \\
\toprule
YAMNET            & .52          & .59          & .57          & .54                                   & .70          & .73          & .74          & .57          & .56          & .61          \\
VGGish            & .57          & .63          & .62          & \textbf{.59}                          & .69          & .75          & .74          & .57          & .58          & .60          \\
\midrule
L3 E 512 L        & .68          & .62          & .65          & .54                                   & .68          & .72          & .66          & .57          & .55          & .57          \\
L3 E 512 M128     & .60          & .63          & .59          & .55                                   & .75          & .62          & .70          & .57          & .58          & .58          \\
L3 E 512 M256     & .63          & .58          & \textbf{.67} & .57                                   & .75          & .70          & .66          & \textbf{.60} & .57          & .61          \\
L3 E 6144 L       & .60          & \textbf{.69} & .59          & .53                                   & .74          & .68          & .73          & .56          & .54          & .55          \\
L3 E 6144 M128    & \textbf{.70} & .67          & .62          & .57                                   & .47          & .70          & .70          & .59          & .55          & .57          \\
L3 E 6144 M256    & .63          & .56          & .58          & .56                                   & .57          & .67          & .69          & .58          & .60          & .57          \\
L3 M 512 L        & .57          & .55          & .65          & .56                                   & .75          & \textbf{.85} & .55          & .58          & .58          & .60          \\
L3 M 512 M128     & .58          & .57          & .58          & .57                                   & .72          & .75          & \textbf{.76} & .58          & \textbf{.61} & .61          \\
L3 M 512 M256     & .56          & .67          & .64          & .54                                   & .71          & .76          & .75          & .57          & \textbf{.61} & \textbf{.62} \\
L3 M 6144 L       & .61          & .56          & \textbf{.67} & .56                                   & .64          & .76          & .61          & .56          & .53          & .59          \\
L3 M 6144 M128    & .58          & .54          & .60          & .56                                   & .75          & .64          & .73          & \textbf{.60} & .58          & .61          \\
L3 M 6144 M256    & .68          & .66          & .63          & .56                                   & \textbf{.81} & .59          & .72          & .57          & .57          & .60  \\
\bottomrule
\end{tabular}

}
\end{table}

Table~\ref{tab:fine_tuning_results} summarizes the performances of the fine-tuned models, highlighting in bold face the best results for each dataset and modality.
With \texttt{Coswara}, both YAMNET and VGGish perform better with the Breath modality, obtaining a PR-AUC score of 0.59 and 0.63, respectively, while OpenL3 performs better with coughing samples and the \texttt{L3 E 6144 M128} configuration, reaching a PR-AUC score of 0.70.
The \texttt{Coughvid} dataset is the only one where OpenL3 performs slightly worse than VGGish.
In fact, while VGGish achieves a score of 0.59, OpenL3 reaches at most 0.57 with 3 different configurations (\texttt{L3 E 512 M256}, \texttt{L3 E 6144 M128}, and \texttt{L3 M 512 M128}), and YAMNET obtains 0.54 of PR-AUC.

By contrast, with \texttt{Cambridge KDD} all the models obtain higher results: YAMNET is able to reach a maximum score of 0.74 with the combination of both coughing and breathing modalities; VGGish obtains a better result (0.75) by using only breath recordings; and, similarly, OpenL3 achieves a score of 0.85 with the \texttt{L3 M 512 L} configuration, based on breathing samples only.
Finally, with the \texttt{Cambridge npj} dataset, all the models perform better by combining the two modalities: YAMNET achieves a PR-AUC score of 0.61, VGGish performs slightly worse, obtaining at most 0.60, while OpenL3 reaches 0.62 with the \texttt{L3 M 512 M256} configuration.

Based on the obtained results, it is worth noting that the fine-tuned models perform considerably worse than their use as feature extractors to input a shallow classifier.
More specifically, compared with the results obtained in the previous set of experiments, the fine-tuned models have lost, on average, 4\%, 6.9\%, and 8.8\% of PR-AUC, respectively for YAMNET, VGGish, and OpenL3.
According to the literature~\cite{raffel2020exploring, PRS19}, the relative performance of the fine-tuning and feature extraction approaches strongly depends on the similarity of the pretraining and target tasks.
Therefore, we can conclude that the considered respiratory sounds do not allow an effective fine-tuning of the analyzed models because they considerably differ from their original training data, which actually include heterogeneous sounds extracted from YouTube videos, including, for example, music, animal sounds, and traffic noises~\cite{45857}.
A possible solution for this issue could be the fine-tuning not only of the final classification layers but also part of the convolutional components dedicated to the feature extraction.
Such an approach could possibly improve the deep representation of the input audio file; however, it typically requires a considerable amount of data that is not currently available in public COVID-19 audio datasets~\cite{9241777}.

\section{Memory footprint}
\label{sec:memory}
In order to investigate the feasibility of a COVID-19 detection system embedded on commercial mobile devices, we compare the memory footprint of the two Transfer Learning approaches evaluated in Section~\ref{sec:experiments}.
In fact, while the computational capabilities of modern smartphones are increasingly comparable with the ones of personal computers and they can even execute complex ML models in just a few milliseconds~\cite{10.1145/3274783.3274840}, the amount of usable memory still represents a limited resource to be optimized.

\begin{figure}[t]
    \begin{subfigure}[h]{0.49\linewidth}
        \centering
        \includegraphics[width=\linewidth]{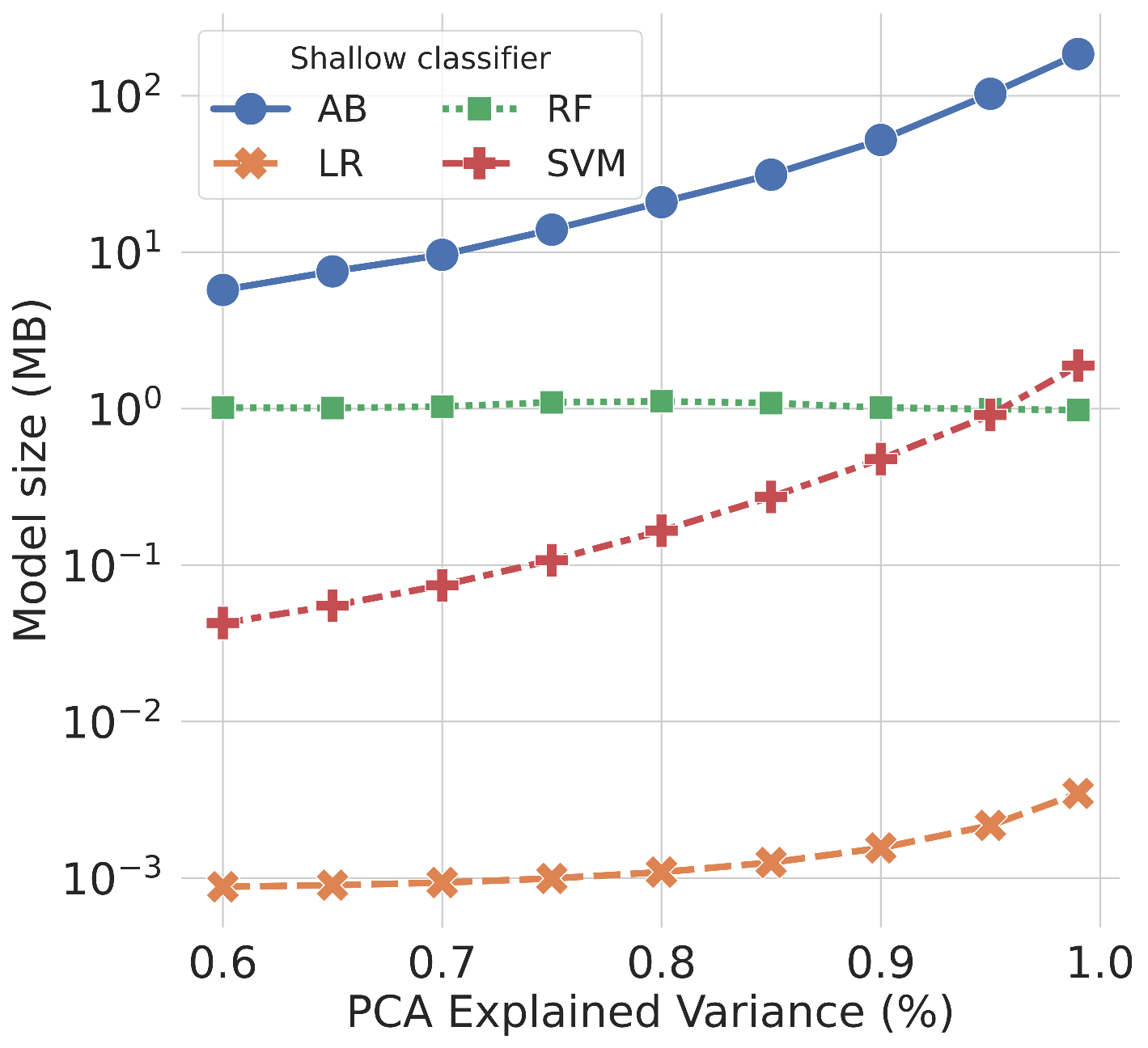}
        \caption{Shallow classifiers}
        \label{fig:shallow_classifier_memory}
    \end{subfigure}
    \hfill
    \begin{subfigure}[h]{0.49\linewidth}
        \includegraphics[width=\linewidth]{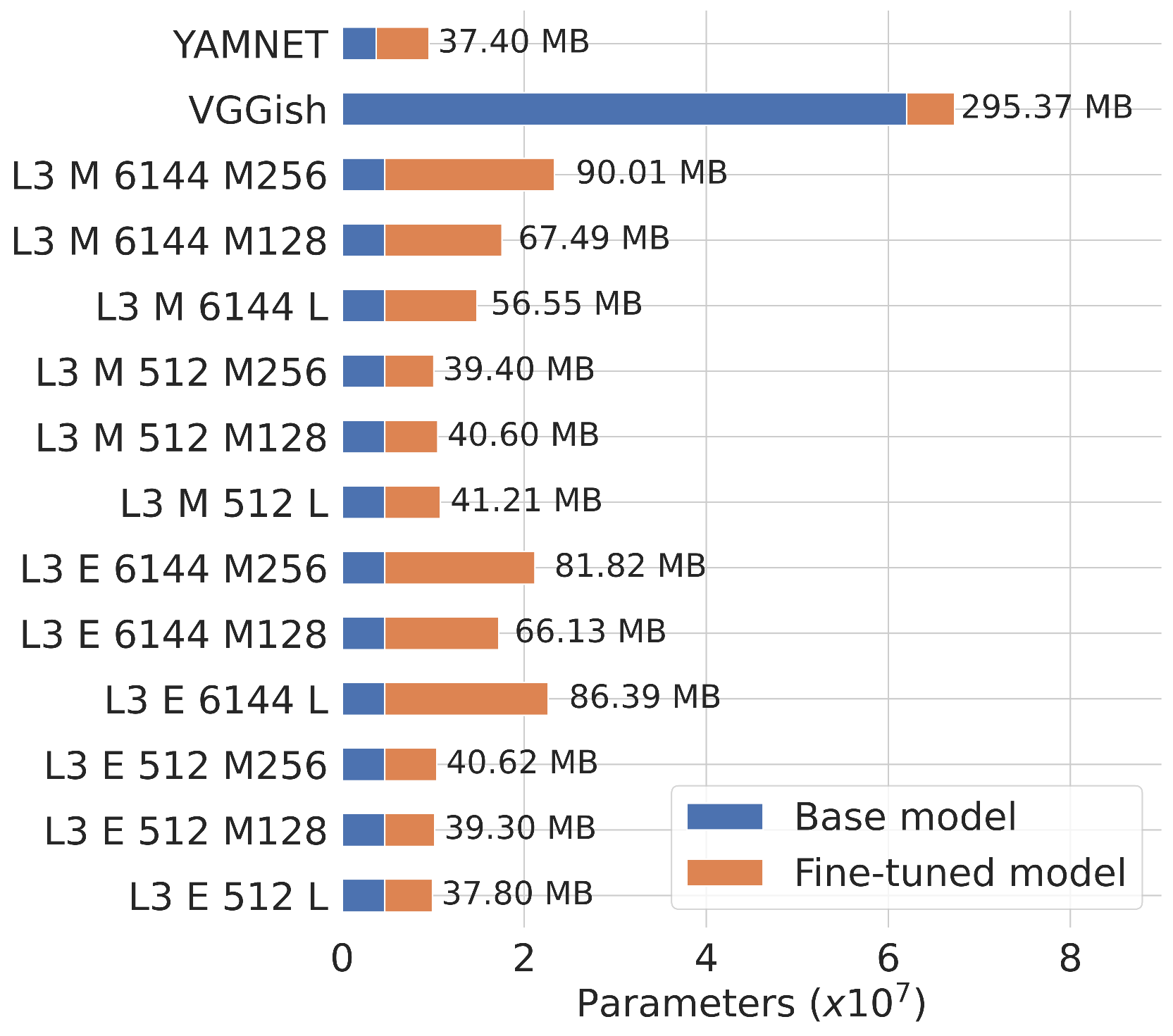}
        \caption{Fine-tuned models}
        \label{fig:fine-tuned_memory}
    \end{subfigure}%
    \caption{Memory footprint of (a) the considered shallow classifiers based on different input sizes (i.e., percentage of PCA explained variance), and (b) the fine-tuned deep learning models. Please, consider that in (a) the overall memory footprint is given by taking into account also the size of the deep embedding models to extract the input features from the raw audio sample.} 
    \label{fig:memory}
\end{figure}

When the deep audio embedding models are used as features extractors, two main components must be considered to estimate the overall memory usage: (i) the size of the deep model used to extract the audio embeddings from the raw input file, and (ii) the size of the shallow model used for the classification phase.
Based on the number of parameters, the sizes of the deep audio embeddings models are the following: YAMNET is the smallest model, with just 3.7M parameters; OpenL3 has 4.7M parameters~\cite{8682475}; while VGGish is the biggest one, with a total of 62M parameters~\cite{7952132}, approximately requiring 16 MB, 288 MB, and 18 MB for YAMNET, VGGish, and OpenL3, respectively.

Moreover, the memory footprint of the classification phase is represented by the size of the employed shallow classifier, which is generally affected by the input data dimension.
Figure~\ref{fig:shallow_classifier_memory} shows the memory size (in MB) of the 4 shallow classifiers considered in our experiments, based on the input size in terms of the percentage of PCA explained variance.
As expected, the size of most of the classifiers increases according to the input dimension, except for Random Forest (RF), whose size remains nearly constant at around 1 MB (between 0.98 and 1.11).
Logistic Regression (LR), the simplest classifier, is also the one with the lowest memory footprint in all the experiments, starting from less than 1 KB (i.e., 922 Bytes), up to 3.64 KB with 99\% of PCA explained variance.
On the other hand, AdaBoost (AB) results to be the most demanding model in terms of memory, with an overall size that ranges from just 5.74 MB with 60\% of PCA, up to 185.05 MB with the full dimension of the input.
Finally, SVM has an intermediate memory footprint among the other classifiers, ranging from 42.6 KB up to 1.88 MB.

On the other hand, when the deep audio models are fine-tuned, the size of the additional fully-connected layers should be considered to estimate the overall memory footprint.
Figure~\ref{fig:fine-tuned_memory} shows the average size of the fine-tuned models, highlighting both the size of the original pre-trained models, and the size of the additional layers for classification.
We can note that, in general, fine-tuning introduces a non-negligible overhead, due to both the number and density of the classification layers considered in our experiments.
More specifically, the overhead introduced by the new layers for YAMNET and VGGish is 5.8M and 5.3M parameters, respectively, which corresponds to an average increment of 33.7 MB for the former model, and 7.37 MB for the latter.
On the other hand, the overhead introduced for OpenL3 strongly depends on the size of the deep audio embeddings: for the configurations that generate a total of 512 embeddings, the fine-tuning introduces, on average, 5.6M parameters (\texttt{+} 21.8 MB); while the configurations with 6144 embeddings have been augmented by 14.8M parameters, which corresponds, on average, to an overhead of 56.73 MB with respect to the base model.

The obtained results clearly show that in both the Transfer Learning approaches, the most demanding component is represented by the pre-trained deep learning model, especially in the case of VGGish, which is the larger model among those considered in our analysis.
Even though the general memory footprint of the considered solutions is manageable on modern smartphones, they can still be prohibitive for mobile devices with less memory resources, such as smartwatches and smart bands.

Therefore, in order to develop novel pervasive m-health applications that can be used as low-cost pre-screening tool for COVID-19, processing the user's data directly on her personal device, the deep learning model should be firstly optimized to be executed on resource-contrained devices.
In this regards, several techniques can be employed to optimize different aspects of deep learning models, including the overall model size~\cite{NIPS2015_ae0eb3ee, han2015deep}.
For example, quantization is a practical and broadly used technique aimed at reducing the size of a pre-trained model by simply lowering the operations precision from 32 bit floats to 16 bit floats or even 8 bit integers~\cite{8927153}, thus reducing the overall model's size by 4x at least, with little degradation in model accuracy, thus enabling its execution on devices with limited computational and memory capabilities.

\section{Conclusions and future work}
\label{sec:conclusions}

In this paper, we investigated the use of different Artificial Intelligence approaches to automatically detect COVID-19 from respiratory sound signals, aimed at developing an effective and low-cost solution to identify new cases, performing the whole data processing directly on the users' mobile devices.
Specifically, we compared the performances of the two most common approaches for audio classifications: the use of handcrafted acoustic features, and pre-trained deep learning models.

In particular, we considered the pre-trained instances of 3 state-of-the-art deep learning models, namely VGGish, YAMNET, and OpenL3, evaluating their efficacy under two transfer learning settings: features extraction and fine-tuning.
In the first case, the deep models are used as simple feature extractors, characterizing respiratory sounds as deep audio embeddings, which are further classified by shallow machine learning classifiers, such as Logistic Regression, and Support Vector Machines.
On the other hand, in the latter approach, the final fully-connected layers of the deep learning models are fine-tuned to adapt the model to the new application domain, thus creating a single model for both data modeling and classification.

Through an extensive evaluation, we compared the performances of the different approaches by using four public datasets and 5-fold nested cross-validation with user-based splits, to avoid biasing the experiments based on specific users or specific train/validation/test splits.
The obtained results clearly show the great advantage of OpenL3 in all the experimental settings, improving the other solutions by 12.3\% in terms of Precision-Recall AUC as features extractor, and by 10\% when the model is fine-tuned.
Moreover, we noted that the fine-tuned models generally perform worse than the feature extractors, with an average drop of 6.6\% of the performances.
Besides, we evaluated the feasibility of the considered solutions of being entirely executed on mobile devices, especially considering their memory constraints.
As imagined, the component with the greatest impact on the overall memory footprint is represented by the deep learning models, especially in the fine-tuning experiments, where additional fully-connected layers are introduced to adapt the model to the new application domain, requiring, on average, between 37 MB and about 300 MB for the most demanding model (i.e., VGGish).

Our future work is threefold.
Firstly, we would like to investigate different techniques to improve the fine-tuning of the deep learning models.
In fact, while we focused on updating only the classification part of the models (i.e., the final fully-connected layers), another possible approach would imply also the partial update of the features extraction component of the model (i.e., the convolutional layers), thus adapting its input representation capabilities to the novel application scenario~\cite{9241777}.
Even though fine-tuning of a larger portion of the model will surely require a non-negligible amount of COVID-19 data, we can still use the combination of the available datasets during the draining process as data augmentation approach.

Secondly, we are planning to develop a prototype mobile application to evaluate the performance of the considered approaches for COVID-19 detection in real-world scenarios.
To this aim, we need to optimize the models in order to be efficiently executed on memory-constrained devices by testing different optimization approaches, including, for example, \emph{quantization}, which reduces the precision of the mathematical operations performed by the neural network, and \emph{pruning}, where redundant connections among hidden units are removed during the training phase~\cite{LIANG2021370}.

Finally, we would like to explore Explainable Artificial Intelligence (XAI) techniques aimed at supporting the predictions produced by the deep learning models.
In particular, since the audio classification models are generally based on Convolutional Neural Networks, we can exploit the same XAI techniques proposed in the literature for visual applications~\cite{VANDERVELDEN2022102470} to describe, for example, which set of frequencies have led the model to classify a given respiratory sound as COVID-positive instead of negative sample, thus providing an additional validation of the classification model.
This is a critical requirement in our reference scenario, since it has been proven that the ability of explaining the output of an AI model increases the faith held in the system, thus leading to a widespread adoption.



\section*{Acknowledgment}

The authors express their gratitude to Prof. Cecilia
Mascolo, Dept. of Computer Science and Technology, University of Cambridge (UK) for sharing the databases of COVID19 sound App described in  ~\cite{exploringcovid2020} and ~\cite{Han2022}.
This work has been partially funded by the European Commission under H2020-INFRAIA-2019-1SoBigData-PlusPlus project. Grant number: 871042.

\bibliographystyle{elsarticle-num}
\bibliography{paper}

\end{document}